\documentclass{article}
\usepackage{iclr2027_conference,times}

\usepackage{amsmath,amsfonts,bm}

\def\eqref#1{equation~\ref{#1}}

\def\1{\bm{1}}

\DeclareMathAlphabet{\mathsfit}{\encodingdefault}{\sfdefault}{m}{sl}
\SetMathAlphabet{\mathsfit}{bold}{\encodingdefault}{\sfdefault}{bx}{n}

\usepackage{amsmath}
\usepackage{amssymb}
\usepackage{graphicx}
\usepackage{booktabs}
\usepackage{multirow}
\usepackage{fvextra}
\usepackage{float}
\usepackage{hyperref}
\usepackage{url}
\usepackage{xcolor}

\title{Memory Internalization via Test-Time Training for Long-Video Understanding}

\author{\parbox[t]{\dimexpr\textwidth-12pt\relax}{\raggedright
\textbf{Haonan Zhao$^{1}$} \quad \textbf{Jiaxuan Chen$^{1}$} \quad \textbf{Qingyang Liu$^{1}$} \quad \textbf{Shuochen Chang$^{1}$}\\
\textbf{Siyu Sun$^{1}$} \quad \textbf{Jiangtong Li$^{2,\dagger}$} \quad \textbf{Li Niu$^{1,\dagger}$}\\[0.4em]
{\normalfont $^{1}$ Shanghai Jiao Tong University \qquad $^{2}$ Tongji University}}}

\iclrfinalcopy
\begin{document}
\raggedbottom

\maketitle
\lhead{Preprint}
\begingroup
\renewcommand{\thefootnote}{\ensuremath{\dagger}}
\footnotetext[1]{Corresponding authors.}
\endgroup

\begin{abstract}
Long-video understanding often requires answering many questions about a single video whose evidence is distributed across extended temporal contexts. Conventional video-language models repeatedly encode frames or retrieve visual evidence for every question. Parameterized video memory offers a write-once alternative, but existing methods rely primarily on synthesized text and fixed adaptation configurations, leaving open where video knowledge should be written and how visual evidence should shape the memory. We introduce \emph{\textbf{M}emory \textbf{I}nternalization via \textbf{T}est-\textbf{T}ime \textbf{T}raining} (\textbf{MIT$^3$}), a per-video adaptation framework that writes each long video into compact, video-specific adaptive parameters for video-free, text-only question answering. Under compact multi-scale supervision, causal tracing measures the causal memory utility of each model layer for video-grounded answering and selects video-specific adaptation locations. Visual memory distillation then trains the text-only memory to reproduce the representation change induced by the relevant video clip, preserving the internal effect of visual evidence beyond textual supervision. Across LVBench, LongVideoBench, and Video-MME, MIT$^3$ achieves data-efficient video-free question answering with substantially fewer synthesized samples than prior parameterized video-memory methods, demonstrating that effective video internalization depends on both where knowledge is written and how visual evidence is distilled into the model.
\end{abstract}

\section{Introduction}
\label{sec:introduction}

Long videos increasingly serve as high-value, persistent information sources, from lecture archives and meeting recordings to egocentric streams and personal media collections.
In these settings, a central task is long-video question answering (QA): a single video must support many questions whose evidence may be brief, recur with subtle changes, or depend on distant events. Modern video-language models~(VLM) typically answer each question by placing sampled frames or compressed visual tokens into the language-model context~\citep{maaz2024video,zhang2023video}.
Sparse context can omit critical evidence, whereas dense context repeatedly incurs visual-token and memory costs, making efficient knowledge reuse for subsequent questions a central challenge for long-video QA.

Existing approaches improve the video-conditioned question-answering path through frame selection, token compression, sparse memories, redundancy reduction, and longer multimodal contexts~\citep{song2024moviechat,chen2025longvila,wang2024retake}.
Retrieval-based methods complement these architectures by preprocessing videos into searchable memories and locating relevant clips for each question~\citep{ren2026videorag}.
These strategies make long inputs more tractable, but their inference path remains video-dependent: each question still requires visual encoding, cached representations, retrieval, or inspection of relevant clips.
When one video must support many questions, as in lecture archives, meetings, surveillance review, or personalized video assistants, repeatedly reconstructing much of the same visual context becomes an avoidable systems bottleneck.

\begin{figure}[t]
    \centering
    \includegraphics[width=\linewidth]{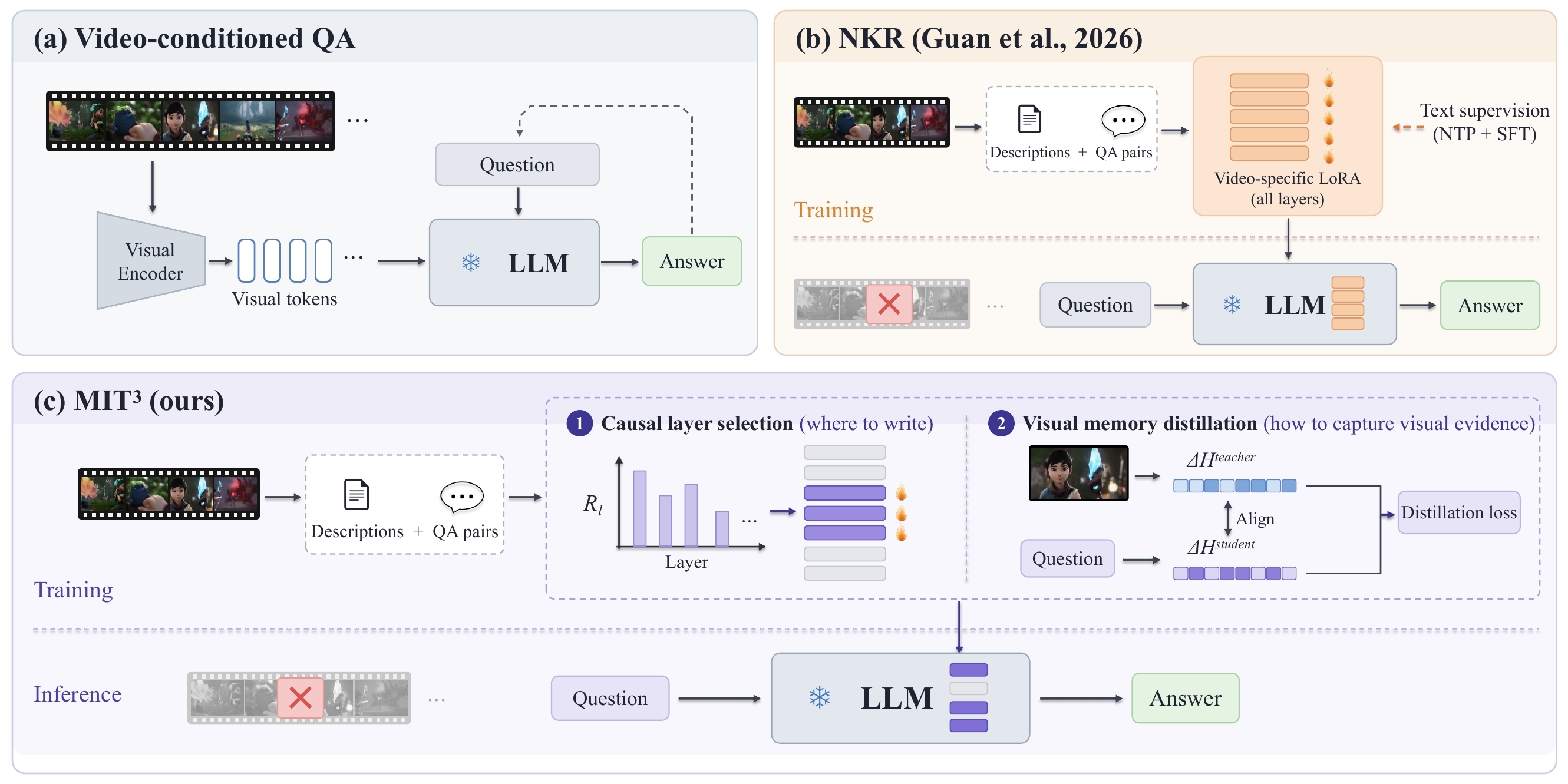}
    \caption{Comparison of video-conditioned question answering, NKR, and MIT$^3$. Conventional video-language models encode or retrieve visual evidence per question. NKR trains a video-specific LoRA adapter on synthesized text using a fixed all-layer configuration. MIT$^3$ instead selects video-specific model layers through causal tracing and distills the visual-evidence-induced representation change into these memory-writing locations, enabling video-free question answering.}
    \label{fig:intro_comparison}
\end{figure}

Video-free inference offers a complementary scaling axis for this setting: it writes a video's knowledge into compact parameters once and answers subsequent questions from text alone.
Neural Knowledge Representation~(NKR)~\citep{guan2026content} instantiates this idea by generating descriptions and question-answer pairs from a video and fitting a video-specific adapter~\citep{hu2021lora}.
Once fitted, the adapter acts as compact parametric memory for text-only question answering, a formulation contrasted with conventional video-dependent inference in Fig.~\ref{fig:intro_comparison}.
Effective internalization therefore poses two knowledge-writing questions.
First, \emph{where should a particular video's knowledge be written?} Transformer layers play distinct computational roles, whereas existing adaptation configurations are typically fixed across videos.
Second, \emph{how should memory capture the effect of visual evidence?} Synthetic descriptions provide factual supervision, while video-conditioned representations directly signal how seeing the video changes internal model computation.

To address these two questions, we propose \emph{\textbf{M}emory \textbf{I}nternalization via \textbf{T}est-\textbf{T}ime \textbf{T}raining} (\textbf{MIT$^3$}), a test-time training framework for video-specific adaptive memory.
To determine \emph{where} to write, MIT$^3$ uses causal tracing to extract the hidden-state change induced by video input, inject this change at each model layer, and score the resulting improvement in answer likelihood.
Layers with the highest causal memory utility are selected as video-specific adaptive locations, where the frozen backbone most effectively converts video evidence into answer-relevant computation.
To determine \emph{how} to write, visual memory distillation uses a frozen video-conditioned teacher to supply the representation change induced by the relevant clip, and trains the text-only adapter to reproduce that change from the question.
With compact multi-scale supervision, these mechanisms store video knowledge at effective computational sites while preserving the internal effect of visual evidence.

To evaluate the effectiveness of MIT$^3$, we conduct experiments on LVBench~\citep{wang2025lvbench}, LongVideoBench~\citep{wu2024longvideobench}, and Video-MME~\citep{fu2025video}, spanning long-video and general video question answering across diverse durations.
On LVBench, MIT$^3$ improves the matched NKR-style baseline by 6.8 points with Qwen3-VL-8B and by 5.7 points with Qwen3-VL-32B, demonstrating the value of causal localization and visual representation matching under identical textual supervision.
The gains extend to LongVideoBench-Long and Video-MME, where MIT$^3$ achieves 63.3 and 66.7 accuracy, respectively, confirming the value of the proposed memory-writing mechanism across extreme-duration and general video understanding settings.
Ablation studies isolate the contributions of memory localization and visual distillation.
Our main contributions are:

\begin{itemize}
\item We study the layer-wise video-memory distribution in the language model and find that layers with high causal memory utility exhibit a concentration pattern across videos.
\item We introduce MIT$^3$, a test-time memory-writing framework that combines causal localization with visual memory distillation for video-free question answering.
\item We demonstrate data-efficient video internalization across LVBench, LongVideoBench, and Video-MME with substantially less synthesized supervision.
\end{itemize}

\section{Related Work}
\label{sec:related_work}

\subsection{Video Understanding}

Video question answering requires models to ground language in visual entities, actions, temporal order, and causal relations.
Early benchmarks progressively expanded this reasoning demand from short animated clips to more complex web videos and causal-temporal questions\citep{tapaswi2016movieqa,lei2018tvqa,xiao2021next}.
Modern video-language models connect a visual encoder to an LLM through learned projection or cross-modal modules and use instruction tuning to support question answering~\citep{maaz2024video,zhang2023video,lin2024video,jin2024chat}.
They answer by integrating sampled frames or visual tokens into the language-model context, making the quality and cost of reasoning depend on the visual information available for each question.

\subsection{Long-Video Context Modeling}

Long-video question answering further requires locating and combining evidence scattered over minutes or hours.
Recent benchmarks evaluate long-range retrieval, temporal reasoning, and event comprehension in this setting~\citep{mangalam2023egoschema,fu2025video,wang2025lvbench,wu2024longvideobench}.
Existing methods extend multimodal context capacity~\citep{chen2025longvila,zhang2024long}, compress observations into sparse memories~\citep{song2024moviechat,he2024ma}, reduce temporal and knowledge redundancy~\citep{wang2024retake,ren2024timechat}, or retrieve relevant evidence from external multimodal indices~\citep{ren2026videorag}.
Although these approaches improve access to long-range evidence, answering each question still requires video tokens, cached representations, or retrieval.
They therefore optimize access to video evidence rather than internalizing a video's knowledge for subsequent text-only question answering.

A complementary direction internalizes a video's knowledge into video-specific parameters, enabling video-free, text-only question answering after one-time per-video adaptation.
NKR~\citep{guan2026content} synthesizes multi-scale descriptions and question-answer pairs and trains a video-specific adapter~\citep{hu2021lora}, demonstrating the feasibility of parameterized video memory.
However, its internalization process primarily relies on synthesized text and adopts a fixed adaptation configuration, without explicitly determining where video knowledge should be written within the backbone or how visual evidence should shape the resulting memory.

\subsection{Test-Time Training}

Test-time training adapts models on test inputs with auxiliary self-supervised objectives to improve robustness under distribution shifts~\citep{sun2020test,liu2021ttt++,gandelsman2022test,zhang2022memo}.
Related fully test-time adaptation methods also update models directly from unlabeled test data, for example through entropy minimization~\citep{wang2020tent,wang2022continual,niu2022efficient,niu2023towards}.
Unlike these methods, MIT$^3$ treats each long video as a test-time memory-writing instance, using the video once to construct adaptive parameters that answer subsequent text-only questions.
This formulation introduces two video-memory-specific requirements beyond standard test-time training: determining where video knowledge should be written and how visual evidence should shape the resulting memory.
MIT$^3$ addresses them through causal tracing for video-specific adaptation locations and visual memory distillation for video-induced representation changes.

\section{Method}
\label{sec:method}

\subsection{Overview}
\label{sec:overview}


Let $V$ be a long video, $Q$ a question about its content, and $A$ the corresponding answer. 
A conventional video-language model $\mathcal{M}$ performs video-conditioned question answering as
\begin{equation}
A \sim p_{\mathcal{M}}(A \mid V,Q),
\end{equation}
requiring $V$, or representations derived from it, whenever a question is asked. 
We instead write the knowledge of $V$ into compact adaptive parameters $\theta_V$ and answer subsequent questions from text:
\begin{equation}
A \sim p_{\mathcal{M}_{\theta_V}}(A \mid Q).
\end{equation}
We instantiate $\theta_V$ as LoRA parameters~\citep{hu2021lora} inserted into the attention sublayers of selected model layers, where each model layer corresponds to one Transformer block.
The backbone remains frozen, and each video is represented by its own lightweight adapter.

\begin{figure}[t]
    \centering
    \includegraphics[width=\linewidth]{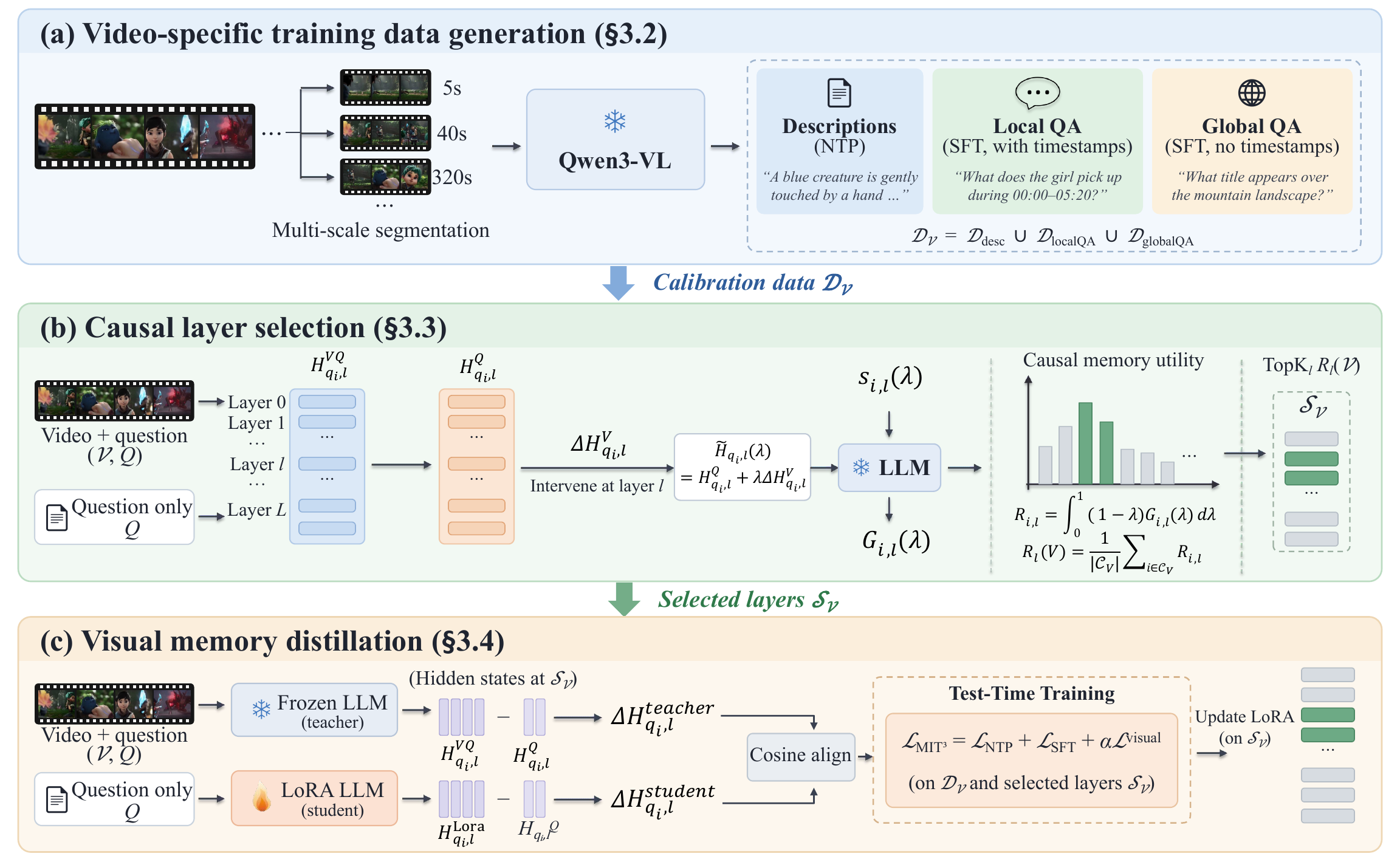}
    \caption{Overview of MIT$^3$.
(a) A VLM constructs multi-scale descriptions, temporally grounded local QA, and global QA to cover video knowledge.
(b) Delta-injection causal tracing estimates each model layer's causal memory utility and selects top-$K$ layers under a fixed adaptation budget.
(c) Visual memory distillation optimizes LoRA modules in selected layers to reproduce normalized video-induced representation changes, transferring visual evidence into text-only memory.}
    \label{fig:overview}
\end{figure}

MIT$^3$ internalizes each video through three stages, shown in Fig.~\ref{fig:overview}. 
First, a VLM constructs compact multi-scale descriptions and local and global question-answer pairs. 
Second, delta-injection causal tracing estimates each model layer's causal memory utility from a small set of temporally grounded local QA pairs and selects top-$K$ layers under a fixed adaptation budget.
Third, LoRA modules in selected layers are optimized with textual supervision and visual memory distillation, which trains the adapter to reproduce normalized video-induced representation changes.
These stages determine what knowledge to write, where to write it, and how visual evidence shapes the memory.

\subsection{Video-Specific Training Data Generation}
\label{sec:data_generation}


The first stage constructs compact, video-specific supervision that preserves both fine-grained evidence and longer temporal structure.
Following NKR~\citep{guan2026content}, we adopt a multi-scale temporal partitioning paradigm. Specifically, our implementation segments $V$ into a temporal pyramid whose window durations double from 5 seconds (i.e., 5, 10, 20, 40, $\ldots$ seconds) up to the full video length.
Short-scale windows are non-overlapping, while longer-scale windows overlap to retain events crossing segment boundaries.
For each window, a VLM generates complementary low-, mid-, and high-level descriptions covering visible details, actions and transitions, and the overall content.
For windows up to 40 seconds, overlapping information across the three levels is fused into one coherent, redundancy-reduced description; for windows of 80 seconds or longer, the three descriptions are retained separately.
Further details on overlap and related configurations are provided in Appendix~\ref{app:training_samples}.


Using the generated descriptions, we construct local and global QA supervision without another video pass.
Local QA pairs are generated from individual window descriptions and cover entities, actions, attributes, temporal order, and clip-level reasoning.
Each local question includes the timestamp or temporal range of its source window, allowing the adapter to encode temporal position and order while identifying the supporting video segment for visual memory distillation in Section~\ref{sec:visual_memory}.
At inference, questions retain the native format of each evaluation benchmark.
To capture video-wide knowledge, we merge recurring entities and selected long-window descriptions into a global memory that summarizes the event timeline, state changes, and major stages of the video.
This global memory conditions global-QA generation and is also included as one next-token prediction example.
Each generation round selects a different group of focus clips and produces $N$ timestamp-free global QA pairs, cycling through the groups to distribute supervision across the video.
These global questions target cross-clip retrieval and reasoning.
The resulting descriptions and QA pairs form the video-specific training set:
\begin{equation}
    \mathcal{D}_V = \mathcal{D}_{\mathrm{desc}} \cup
    \mathcal{D}_{\mathrm{localQA}} \cup
    \mathcal{D}_{\mathrm{globalQA}}.
\end{equation}
During adapter optimization, $\mathcal{D}_{\mathrm{desc}}$ provides next-token prediction supervision, while $\mathcal{D}_{\mathrm{localQA}}$ and $\mathcal{D}_{\mathrm{globalQA}}$ provide fine-tuning examples.
Temporally grounded local QA pairs associate each example with its supporting video segment for visual memory distillation, while description and global-QA examples remain text-only.
Our description and entity construction follow the multi-level generation design of NKR~\citep{guan2026content}, while local-QA generation adapts its evidence format to open-ended answering and the global-memory and full-video QA components provide video-wide supervision introduced by MIT$^3$.
Appendix~\ref{app:prompts} provides the corresponding construction details.

\subsection{Causal Localization of Video Memory}
\label{sec:layer_selection}


The second stage localizes where to write knowledge from $V$ in the frozen backbone.
Adapting every model layer distributes memory parameters throughout the backbone, while a fixed subset assumes shared storage locations across videos.
MIT$^3$ instead uses delta-injection causal tracing to measure each model layer's video-specific causal memory utility.
For a candidate layer $l$, we ask: \emph{if the hidden-state change induced by the video is injected during question-only inference, how strongly can the remaining computation convert it into a higher likelihood for the paired answer?}
The resulting answer gain ranks model layers by utility as video-specific memory-writing locations.


\textbf{Video-induced representations.}
For each calibration triple $(V_i,Q_i,A_i)$, we run the frozen model with the relevant video segment and question, then with the question alone.
Let $q_i$ denote the aligned positions of identical question tokens in both forward passes.
Let $H^{VQ}_{q_i,l}$ and $H^{Q}_{q_i,l}$ denote their hidden states at model layer $l$ under video-conditioned and question-only inference, respectively.
We define the video-induced representation change on these aligned question tokens as
\begin{equation}
    \Delta H_{q_i,l}^{V}
    = H^{VQ}_{q_i,l} - H^{Q}_{q_i,l}.
    \label{eq:video_delta}
\end{equation}
This delta captures how observing the relevant video segment changes the same question representations and provides the visual signal shared by causal localization and visual memory distillation.


\textbf{Hidden-state intervention.}
We next measure how effectively the computation after model layer $l$ can use the video-induced delta.
For an intervention strength $\lambda\in[0,1]$, we construct
\begin{equation}
    \widetilde{H}_{q_i,l}(\lambda)
    = H^{Q}_{q_i,l}
    + \lambda \Delta H_{q_i,l}^{V},
    \qquad \lambda \in [0,1],
    \label{eq:interpolation}
\end{equation}
where small $\lambda$ values test whether the remaining computation can exploit a limited video-induced change, while $\lambda=1$ injects the complete question-token delta at layer $l$.
We replace the question-token states at layer $l$ with $\widetilde{H}_{q_i,l}(\lambda)$ and continue question-only inference through subsequent model layers.
Let $s_i^Q$ and $s_{i,l}(\lambda)$ denote the answer log-likelihood before and after intervention:
\begin{equation}
        s_i^Q = \log p_{\mathcal{M}}(A_i \mid Q_i), \quad\quad\quad
        s_{i,l}(\lambda)= \log p_{\mathcal{M}}(A_i \mid \widetilde{H}_{q_i,l}(\lambda)).
\end{equation}
The intervention gain is
$G_{i,l}(\lambda)= s_{i,l}(\lambda)-s_i^Q.\label{eq:answer_gain}$
A larger gain indicates that computation after layer $l$ more effectively converts the injected video-induced change into evidence for the paired answer.



\textbf{Budget-averaged causal memory utility.}
The intervention curve $G_{i,l}(\lambda)$ characterizes the causal memory utility of model layer $l$ as increasing amounts of the video-induced delta are injected.
A compact adapter benefits from layers where even a limited representation change produces answer-relevant gains, rather than only becoming useful near full intervention.
We therefore introduce $b\in[0,1]$ as the maximum intervention strength under consideration.
For each $b$, the prefix integral $\int_0^b G_{i,l}(\lambda)\,d\lambda$ measures the cumulative answer gain available within that intervention budget.
Averaging these prefix gains over all budgets gives
\begin{equation}
    \begin{aligned}
    R_{i,l}
    = \int_{0}^{1}\int_{0}^{b}
    G_{i,l}(\lambda)\,d\lambda\,db
    = \int_{0}^{1}(1-\lambda)
    G_{i,l}(\lambda)\,d\lambda.
    \end{aligned}
    \label{eq:memory_utility}
\end{equation}
The equivalent weighted form emphasizes gains obtained from smaller interventions while retaining the utility accumulated across the full intervention trajectory.
A high $R_{i,l}$ therefore identifies a layer that responds to limited video-induced changes and continues to provide answer-relevant utility as more visual information is injected.



For each video $V$, let $\mathcal{C}_V$ denote a calibration set of temporally grounded triples $(V_i,Q_i,A_i)$ sampled across temporal scales.
Averaging the example-level scores gives the video-level causal memory utility of model layer $l$:
\begin{equation}
    R_l(V)
    = \frac{1}{|\mathcal{C}_V|}
    \sum_{i\in\mathcal{C}_V} R_{i,l}.
    \label{eq:video_layer_score}
\end{equation}
Under the fixed adaptation budget $K$, we select the layers with the highest video-level utilities:
\begin{equation}
    \mathcal{S}_V
    = \operatorname*{TopK}_{l}\;R_l(V).
    \label{eq:topk_layers}
\end{equation}
The utility scores are computed once with the frozen model and without gradients before adapter optimization.
LoRA modules are inserted only into the attention sublayers of model layers in $\mathcal{S}_V$, while the remaining layers and all backbone parameters stay frozen.
This converts adapter placement from a fixed architectural configuration into a video-specific memory-writing decision.

\subsection{Visual Memory Distillation}
\label{sec:visual_memory}


The third stage writes visual evidence into the selected memory locations by distilling how video conditioning redirects question representations.
Synthesized descriptions provide explicit semantic supervision, while video-conditioned hidden states supply complementary signals about appearance, spatial configuration, and transient events.
For each temporally grounded local QA example, the frozen model processes the relevant video segment and question as teacher, while the adapter-enabled student receives the same question without video input.
Visual memory distillation trains the normalized adapter-induced changes to reproduce the normalized video-induced representation changes at the selected model layers $\mathcal{S}_V$.
This transfers the internal computational effect of the relevant video segment into text-only memory without retaining frames or visual tokens.


\textbf{Teacher and student deltas.}
At each selected model layer $l\in\mathcal{S}_V$, the frozen teacher provides the video-induced target defined in Equation~\ref{eq:video_delta}:
\begin{equation}
    \Delta H_{q_i,l}^{\mathrm{teacher}}
    = H_{q_i,l}^{VQ}-H_{q_i,l}^{Q}
    = \Delta H_{q_i,l}^{V}.
    \label{eq:teacher_delta}
\end{equation}
The text-only student processes $Q_i$ with the video-specific adapter enabled.
Let $H_{q_i,l}^{\mathrm{LoRA}}$ denote its hidden states at the same aligned question-token positions.
The adapter-induced change is
\begin{equation}
    \Delta H_{q_i,l}^{\mathrm{student}}
    = H_{q_i,l}^{\mathrm{LoRA}}
    - H_{q_i,l}^{Q}.
    \label{eq:student_delta}
\end{equation}
Both deltas are measured relative to the same question-only state $H_{q_i,l}^{Q}$.
This shared reference isolates the transformations contributed by the video and the adapter, making their representation changes directly comparable.


\textbf{Representation alignment.}
We align the teacher and student deltas using cosine distance.
For model layer $l\in\mathcal{S}_V$, the token-averaged loss is
\begin{equation}
    \mathcal{L}_{i,l}^{\mathrm{visual}}
    = \frac{1}{|q_i|}
    \sum_{t\in q_i}
    \left(
    1-
    \frac{
    \left\langle
    \Delta h_{i,l,t}^{\mathrm{student}},
    \Delta h_{i,l,t}^{\mathrm{teacher}}
    \right\rangle
    }{
    \left\|
    \Delta h_{i,l,t}^{\mathrm{student}}
    \right\|_2
    \left\|
    \Delta h_{i,l,t}^{\mathrm{teacher}}
    \right\|_2
    + \epsilon
    }
    \right),
    \label{eq:visual_layer_loss}
\end{equation}
where $\epsilon$ provides numerical stability.
The per-example visual loss averages this reproduction objective over the selected model layers:
\begin{equation}
    \mathcal{L}_{i}^{\mathrm{visual}}
    = \frac{1}{|\mathcal{S}_V|}
    \sum_{l\in\mathcal{S}_V}
    \mathcal{L}_{i,l}^{\mathrm{visual}}.
    \label{eq:visual_loss}
\end{equation}
This loss is evaluated for the temporally grounded local QA examples used for visual distillation, while description and global-QA examples provide text-only NTP and SFT supervision.




\subsection{Test-Time Memory Optimization}
\label{sec:training_objective}

For each video, we freeze the base VLM and optimize only LoRA parameters in attention sublayers of the causally selected model layers $\mathcal{S}_V$.
Let $\mathcal{D}_{\mathrm{visual}}\subseteq\mathcal{D}_{\mathrm{localQA}}$ denote the temporally grounded local QA examples used for visual memory distillation.
Their sample-averaged visual loss is
\begin{equation}
    \mathcal{L}_{\mathrm{visual}}
    = \frac{1}{|\mathcal{D}_{\mathrm{visual}}|}
    \sum_{i\in\mathcal{D}_{\mathrm{visual}}}
    \mathcal{L}_{i}^{\mathrm{visual}}.
    \label{eq:visual_dataset_loss}
\end{equation}
The complete test-time memory optimization objective is
\begin{equation}
    \mathcal{L}_{\mathrm{MIT}^3}
    = \mathcal{L}_{\mathrm{NTP}}
    + \mathcal{L}_{\mathrm{SFT}}
    + \alpha\mathcal{L}_{\mathrm{visual}},
    \label{eq:full_objective}
\end{equation}
where each term is averaged over its corresponding training subset and $\alpha$ controls the visual-distillation weight.
After this one-time optimization, the resulting adapter $\theta_V$ serves as the video-specific memory.
At inference, the frozen backbone receives only the question $Q$ with $\theta_V$ enabled, requiring no video input, visual tokens, or retrieval.
The same adapter can answer subsequent questions about $V$ without reprocessing the video.

\begin{table}[t]
    \caption{Accuracy and inference efficiency on LVBench and LongVideoBench. Time/query denotes inference time per question. $\dagger$ denotes controlled NKR-style baseline using the same data as MIT$^3$.}
    \label{tab:main_results}
    \centering
    \begin{tabular*}{\linewidth}{@{\extracolsep{\fill}}llcccccc@{}}
        \toprule
        \multirow{2}{*}{Method} & \multirow{2}{*}{Input} &
        \multirow{2}{*}{\shortstack{Time/query\\(s)}} &
        \multirow{2}{*}{\shortstack{Mem.\\(GB)}} &
        \multicolumn{2}{c}{LVBench} & \multicolumn{2}{c}{LongVideoBench} \\
        \cmidrule(lr){5-6}\cmidrule(lr){7-8}
        & & & & All & $>1$h & All & Long \\
        \midrule
        VideoLLaMA3-7B & $<768$ fr. & 64.5 & 1.2 & 45.3 & -- & -- & -- \\
        AdaReTaKe-7B & $<2048$ fr. & 45 & 0.9 & 51.2 & 48.4 & -- & 55.5 \\
        VideoRAG & N/A & 90 & 0.5 & 49.2 & 47.8 & -- & -- \\
        VCA & N/A & -- & -- & 41.3 & -- & -- & -- \\
        DVD-o3 & N/A & 180 & 0.3 & 74.2 & -- & -- & -- \\
        NKR-8B & N/A & 0.25 & 0 & 39.5 & -- & -- & -- \\
        NKR-14B & N/A & 0.33 & 0 & 48.8 & 48.6 & -- & 52.9 \\
        \midrule
        \multicolumn{8}{l}{\textit{Qwen3-VL-8B backbone, with 2 fps}} \\
        \quad Direct & $<768$ fr. & 31.1 & 15.1 & 45.6 & 42.1 & 60.2 & 48.0 \\
        \quad NKR-style$^{\dagger}$ & N/A & 0.25 & 0 & 36.7 & 36.5 & 54.5 & 43.3 \\
        \quad MIT$^3$ (Ours) & N/A & 0.25 & 0 & 43.5 & 43.2 & 58.5 & 47.1 \\
        \midrule
        \multicolumn{8}{l}{\textit{Qwen3-VL-32B backbone, with 2 fps}} \\
        \quad Direct & $<768$ fr. & 103.4 & 26.9 & 52.5 & 48.1 & 65.7 & 55.1 \\
        \quad NKR-style$^{\dagger}$ & N/A & 0.46 & 0 & 44.9 & 45.3 & 57.4 & 47.3 \\
        \quad MIT$^3$ (Ours) & N/A & 0.46 & 0 & 50.6 & \textbf{50.1} & 63.3 & \textbf{55.9} \\
        \bottomrule
    \end{tabular*}
\end{table}

\section{Experiments}
\label{sec:experiments}

\subsection{Experimental Setup}
\label{sec:experimental_setup}

\paragraph{Datasets and Baselines}\mbox{}\\
We evaluate MIT$^3$ on LVBench~\citep{wang2025lvbench}, LongVideoBench~\citep{wu2024longvideobench}, and Video-MME~\citep{fu2025video}. For LVBench and LongVideoBench, we report results on both the full benchmarks and their long-video subsets, excluding subtitle and audio inputs. Baselines include published long-video methods~\citep{zhang2025videollama,wang2025adaretake,ren2026videorag,yang2025vca,zhang2026deep,guan2026content}, direct Qwen3-VL-8B/32B inference~\citep{bai2025qwen3}, and a controlled NKR-style baseline. Since NKR's per-video training data are unavailable, we reproduce its data-generation procedure and train both NKR-style and MIT$^3$ on the same generated supervision. The NKR-style baseline optimizes only $\mathcal{L}_{\mathrm{NTP}}+\mathcal{L}_{\mathrm{SFT}}$, while MIT$^3$ additionally uses video-specific layer selection and visual memory distillation.

\paragraph{Implementation Details}
Following video-NKR~\citep{guan2026content}, we use LoRA with rank 256 and scaling factor 512, selecting the top 18 and 32 layers for Qwen3-VL-8B and Qwen3-VL-32B, respectively. For a one-hour video, our pipeline generates approximately 5.5k training examples, roughly one-eighth of those used by NKR. We also verify that no evaluation questions appear in adapter training or causal-localization calibration. Full dataset, data-generation, and optimization details are provided in Appendix~\ref{app:implementation_details}.

\subsection{Main Results}
\label{sec:main_results}

Table~\ref{tab:main_results} compares MIT$^3$ with published long-video systems, direct video-conditioned inference, and the matched NKR-style baseline. 
Under identical video-specific supervision, MIT$^3$ consistently outperforms the NKR-style baseline across both backbone sizes and all four evaluation settings, demonstrating the effectiveness of video-specific layer selection and visual memory distillation for parameterized video memory. 
Despite performing video-free, text-only inference after one-time internalization, MIT$^3$ remains close to direct video-conditioned inference on the full benchmarks, while requiring no video frames, visual tokens, retrieval, or video-dependent KV cache at query time. 
More importantly, with the 32B backbone, MIT$^3$ reaches 50.1 on LVBench $>1$h and 55.9 on LongVideoBench-Long, surpassing direct video-conditioned inference on both long-video subsets. 
Together with the small variation between the full LVBench and its $>1$h subset, these results highlight the effectiveness and stability of MIT$^3$ for long-video understanding. 
This makes MIT$^3$ particularly suitable for scenarios that repeatedly revisit long videos, such as archived surveillance review, recorded meetings, and lecture or instructional video analysis. 
We further analyze the resource advantages of MIT$^3$ over alternative inference paradigms in Appendix~\ref{app:deployment_cost}.

\subsection{Layer-Wise Distribution}
\label{sec:layer_distribution}

\begin{figure}[t]
    \centering
    \IfFileExists{figures/layer_distribution_8b.pdf}{
        \includegraphics[width=\linewidth]{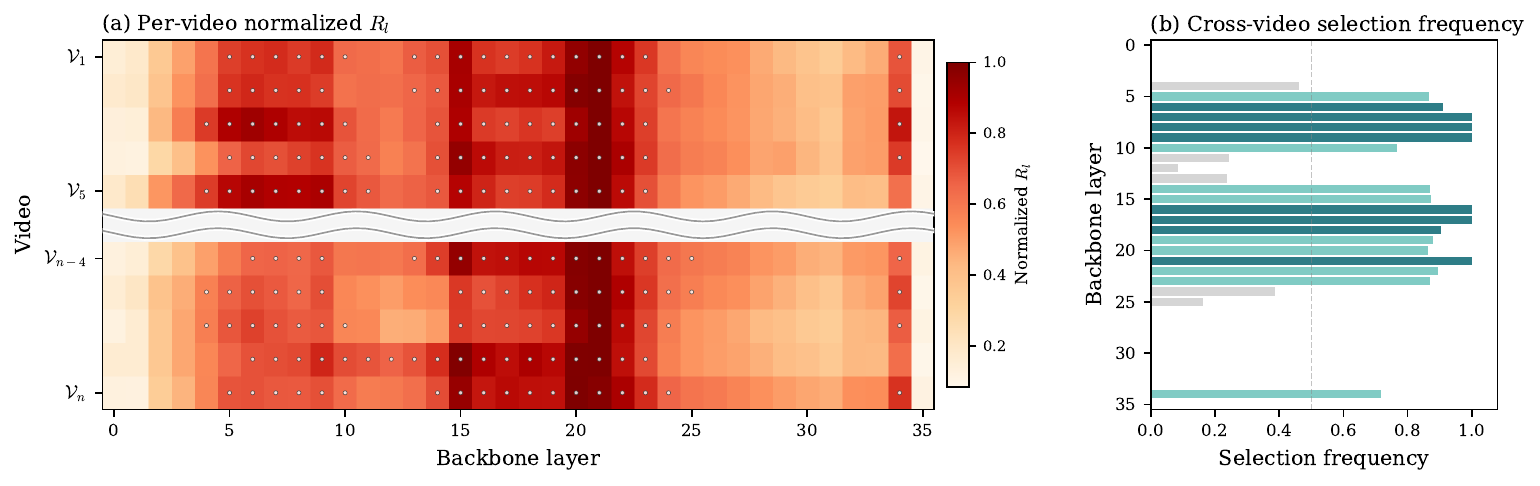}
    }{
        \fbox{\begin{minipage}[c][1.55in][c]{0.94\linewidth}
            \centering
            \textbf{Qwen3-VL-8B layer-distribution visualization}\par\medskip
            \begin{minipage}[t]{0.47\linewidth}
                \centering
                \textbf{(a) Video--layer heatmap}\par
                Rows: videos; columns: backbone layers.\par
                Color: normalized $R_l(V)$; markers: top-18 layers.
            \end{minipage}\hfill
            \begin{minipage}[t]{0.47\linewidth}
                \centering
                \textbf{(b) Cross-video selection frequency}\par
                Horizontal axis: backbone layer.\par
                Vertical axis: fraction of videos selecting the layer.
            \end{minipage}
        \end{minipage}}
    }
    \caption{Layer-wise causal utility and top-18 selection frequency for Qwen3-VL-8B on LVBench.}
    \label{fig:layer_distribution_8b}
\end{figure}

We analyze the layer-wise causal utilities of Qwen3-VL-8B on LVBench to examine whether video-specific localization selects arbitrary layers or reveals a consistent organization of video memory. 
For each video, we normalize $R_l(V)$ across layers and identify the top-18 layers selected for adaptation. Figure~\ref{fig:layer_distribution_8b}(a) shows the resulting video-layer heatmap, while Figure~\ref{fig:layer_distribution_8b}(b) reports the selection frequency of each layer across videos.

The selected layers show substantial cross-video agreement. 
A compact group of layers is repeatedly assigned high causal memory utility, forming a shared memory-writing core that accounts for \textbf{65\%} of all top-18 selections. 
At the same time, a smaller set of selected layers varies across videos, introducing content-dependent refinements around this common structure. 
These results suggest that causal localization captures both a consistent cross-video organization of memory and video-specific adaptation needs, rather than reducing to either a fixed layer subset or arbitrary per-video selection. 
The corresponding analysis for Qwen3-VL-32B is provided in Appendix~\ref{app:layer_distribution_32b}.

\subsection{Ablation Study}
\label{sec:ablation}

\paragraph{Component ablations.}
Table~\ref{tab:ablation} isolates the contributions of layer selection and visual memory distillation on LVBench using Qwen3-VL-8B. 
All component ablations apply LoRA to the attention projections and compare all-layer, random-$K$, and causal top-$K$ placement with or without the visual loss. 
We further examine the choice of LoRA target by applying the complete MIT$^3$ configuration to either the attention projections or the FFN while keeping all other settings fixed. 
The two choices yield comparable performance, we therefore follow the NKR configuration and use attention-targeted LoRA in all main experiments.

\begin{table}[t]
    \centering
    \caption{Component and LoRA-target ablations on LVBench with Qwen3-VL-8B.}
    \label{tab:ablation}
    \resizebox{0.7\linewidth}{!}{%
    \begin{tabular}{lcccc}
        \toprule
        Configuration & Layer Selection & LoRA Target & Visual Loss & Accuracy (\%) \\
        \midrule
        Baseline & All & Attention & -- & 36.7 \\
        Random-$K$ & Random $K$ & Attention & -- & 34.1 \\
        All + Visual & All & Attention & \checkmark & 41.2 \\
        Causal Top-$K$ & Causal Top-$K$ & Attention & -- & 39.2 \\
        MIT$^3$ & Causal Top-$K$ & Attention & \checkmark & 43.5 \\
        MIT$^3$ (FFN) & Causal Top-$K$ & FFN & \checkmark & 43.9\\
        \bottomrule
    \end{tabular}%
    }
\end{table}

\paragraph{Data efficiency.}
\label{sec:data_efficiency}

\begin{figure}[t]
    \centering
    \IfFileExists{figures/data_efficiency.pdf}{
        \includegraphics[width=0.8\linewidth]{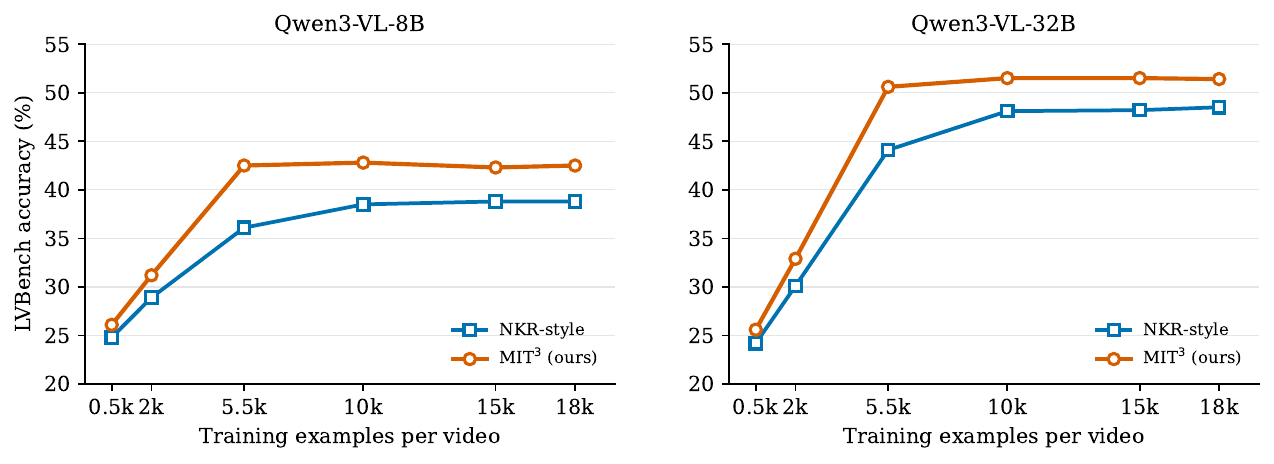}
    }{
        \fbox{\parbox[c][1.35in][c]{0.68\linewidth}{\centering
        Data-efficiency plot placeholder\\[3pt]
        Accuracy vs. training examples (0.5k, 2k, 5.5k, 10k, 15k, 18k)\\
        NKR-style and MIT$^3$ with 8B and 32B backbones}}
    }
    \caption{Data efficiency on LVBench under matched per-video supervision budgets.}
    \label{fig:data_efficiency}
\end{figure}

We further compare NKR-style and MIT$^3$ under matched supervision budgets using Qwen3-VL-8B and Qwen3-VL-32B. 
The budgets in Fig.~\ref{fig:data_efficiency}, 0.5k, 2k, 5.5k, 10k, 15k, and 18k examples per video, are estimated from a one-hour video, with 5.5k corresponding to our standard configuration. 
MIT$^3$ consistently outperforms the matched NKR-style baseline across all supervision budgets and both backbone sizes, indicating that the proposed memory-writing strategy improves the utilization of synthesized supervision rather than relying on a larger training set. 
Notably, the performance of MIT$^3$ approaches saturation once the supervision reaches a moderate scale, and further increasing the number of generated examples yields only marginal gains beyond the 5.5k setting. 
This trend suggests that the benefit of MIT$^3$ is driven primarily by how video knowledge is localized and distilled into the model, rather than by continued scaling of synthetic supervision. 
In other words, causal localization and visual memory distillation allow the model to extract more useful video-specific knowledge from a compact set of training examples.

\section{Conclusion}

We introduced MIT$^3$, a per-video test-time training framework that converts each long video into compact adaptive memory for write-once, query-many understanding. By answering subsequent questions through text-only inference, MIT$^3$ eliminates repeated visual encoding or retrieval while retaining video-specific knowledge in lightweight parameters. Causal localization identifies where this knowledge can be written most effectively, and visual memory distillation trains the adapter to reproduce the representation change induced by visual evidence. Across LVBench, LongVideoBench, and Video-MME, MIT$^3$ consistently improves controlled text-only internalization using roughly one-eighth of NKR's synthesized supervision. Its per-video state is $45.6\times$ smaller than persistent KV reuse with Qwen3-VL-32B, enabling scalable access to growing videos.

\bibliography{arxiv}
\bibliographystyle{iclr2027_conference}

\clearpage
\appendix

\section{Training Data Construction and Examples}
\label{app:training_samples}

This section provides additional implementation details for the multi-scale supervision introduced in Section~\ref{sec:data_generation}, together with representative training examples. To balance temporal coverage and redundancy, the 5-, 10-, and 20-second windows are non-overlapping, while windows of 40 seconds or longer use 50\% overlap with a stride equal to half the window length. For windows up to 40 seconds, overlapping content from the low-, mid-, and high-level descriptions is fused into a single coherent, redundancy-reduced training example; for windows of 80 seconds or longer, the three descriptions are retained as separate examples. Table~\ref{tab:training_data_construction} summarizes the resulting configuration. Figure~\ref{fig:training_examples} illustrates the corresponding training formats using a 5-second fused-description example, an 80-second separate-description example, and representative local and global QA pairs. Ellipses indicate omitted content for compact presentation.

\begin{table}[H]
    \caption{Per-video multi-scale training-data construction. ``Merged'' indicates that overlapping information from the three description levels is fused into one coherent, redundancy-reduced training example.}
    \label{tab:training_data_construction}
    \centering
    \begin{tabular}{lcccc}
        \toprule
        Window & Stride & Overlap & Descriptions & Local QA \\
        \midrule
        5 s   & 5 s   & 0\%  & Merged   & 1 \\
        10 s  & 10 s  & 0\%  & Merged   & 1 \\
        20 s  & 20 s  & 0\%  & Merged   & 3 \\
        40 s  & 20 s  & 50\% & Merged   & 5 \\
        80 s  & 40 s  & 50\% & Separate & 5 \\
        $\geq$160 s & Half-window & 50\% & Separate & 10 \\
        \bottomrule
    \end{tabular}
\end{table}

\begin{figure}[H]
\centering
\fbox{
\begin{minipage}{0.96\linewidth}
\small

\textbf{Short window (5 seconds): fused description example}\\
\textit{Time range: 00:00:00--00:00:05}\\
\textbf{Fused example:}
White serif text on a dark background explains that unique wooden masks are carved to honor the dead and placed in sacred shrines, where they gradually return to dust and symbolize the spirit's peaceful journey into the next life. Subtle blue light orbs drift in the background \ldots

\medskip
\hrule
\medskip

\textbf{Long window (80 seconds): separate description examples}\\
\textit{Time range: 00:00:00--00:01:20}\\
\textbf{[High]} A young girl and two small green-haired children interact in a mystical forest \ldots\\
\textbf{[Mid]} The video opens with text about wooden masks and spirit guides; the girl then observes the children nearby \ldots\\
\textbf{[Low]} White text appears on screen; the girl has dark hair and wears a blue top, while the children have green hair with colorful ornaments \ldots

\medskip
\hrule
\medskip

\textbf{Local QA}\\
\textbf{Q:} From 00:00:00 to 00:02:40, what does the girl do while interacting with the creatures?\\
\textbf{A:} She moves through the forest and interacts closely with the two creatures \ldots

\medskip
\hrule
\medskip

\textbf{Global QA}\\
\textbf{Q:} What does the text in the video say about the purpose of the wooden masks?\\
\textbf{A:} They are carved to honor the dead and symbolize the spirit's peaceful journey \ldots

\end{minipage}
}

\caption{
Representative training examples for the multi-scale supervision pipeline.
For short windows ($\leq 40$ s), overlapping information from the high-, mid-, and low-level descriptions is fused into one coherent, redundancy-reduced training example, whereas for long windows ($\geq 80$ s), the descriptions are retained as three separate examples.
Local QA pairs include explicit temporal grounding, while global QA pairs are timestamp-free.
Ellipses denote omitted content.
}
\label{fig:training_examples}
\end{figure}

\section{Data Generation Prompts}
\label{app:prompts}

This appendix documents the prompts used for data generation in MIT$^3$. We provide the newly introduced prompts in full, while components inherited from or adapted from NKR~\citep{guan2026content} are summarized at the level of their structure and modifications.

\subsection{NKR-Based Components}

Our data-generation pipeline retains several components from NKR~\citep{guan2026content}. Specifically, we preserve its three-level captioning format within a single model call and its two-stage entity-processing procedure. For local clip-level QA generation, we follow the same caption-and-entity evidence organization, but adapt the task from multiple-choice to open-ended QA and require each question to be temporally grounded.

\begin{figure}[t]
\centering
\fcolorbox{black!35}{black!3}{\begin{minipage}{0.92\linewidth}
\raggedright
\textbf{Multi-level clip-description prompt}

\medskip
\textbf{Instructions.} You are a professional video annotator. Based on the input video clip, generate three levels of natural, fluent, and descriptive captions in English.

\medskip
\textbf{Output format.} Return a JSON object with the fields
\texttt{caption\_high\_level}, \texttt{caption\_mid\_level}, and
\texttt{caption\_low\_level}.

\medskip
\textbf{Guidelines.}
\begin{itemize}
\item Capture temporal dynamics, including movements, actions, and transitions.
\item Describe scene evolution factually.
\item Translate factual entities into English and ignore irrelevant expressions.
\end{itemize}

\textbf{Caption levels.}
\begin{itemize}
\item \textbf{Level 1:} a short summary of the main subject.
\item \textbf{Level 2:} a richer description of entities, actions, and context.
\item \textbf{Level 3:} a comprehensive narrative detailing attributes, spatial relations, and temporal evolution.
\end{itemize}
\end{minipage}}
\caption{Prompt for multi-level clip descriptions, adapted from NKR.}
\label{fig:prompt_descriptions}
\end{figure}

\begin{figure}[t]
\centering
\fcolorbox{black!35}{black!3}{\begin{minipage}{0.92\linewidth}
\textbf{Stage 1: clip-level entity extraction}

\medskip
\textbf{Instructions.} You are a professional video entity extractor. Analyze the input video clip and detect all entities. For each entity, provide its name, appearance, identity, and first-seen timestamp within this clip. Avoid duplicates and return a JSON object.

\medskip
\hrule
\medskip
\textbf{Stage 2: video-level entity merging}

\medskip
\textbf{Instructions.} You are a professional assistant for merging entities. Given entity registries from different clips of the same video, merge duplicate entities, retain the earliest first-seen time, and create a \texttt{timestamps} field containing all occurrence times.
\end{minipage}}
\caption{NKR-style prompts for clip-level entity extraction and video-level merging.}
\label{fig:prompt_entities}
\end{figure}

\begin{figure}[t]
\centering
\fcolorbox{black!35}{black!3}{\begin{minipage}{0.92\linewidth}
\textbf{Local open-ended QA prompt}

\medskip
\textbf{Instructions.} You are a careful dataset constructor that converts one video clip into high-quality open-ended QA pairs. Given multi-level captions and entities for a video clip, produce diverse QA pairs covering multiple scales and aspects.

\medskip
\textbf{Question composition.}
\begin{itemize}
\item Cover high-level gist, mid-level attributes and relations, and low-level fine-grained details.
\item Questions may concern a single moment, an action sequence, or multiple events.
\end{itemize}

\textbf{Timestamp rules.}
\begin{itemize}
\item Each timestamp must lie within the clip's time range.
\item Use floating-point seconds and prefer ranges for sustained actions.
\item Include the timestamp or temporal range directly in the question.
\end{itemize}

\textbf{Workflow.}
\begin{enumerate}
\item Parse the captions into atomic facts.
\item Fuse them into a contradiction-free fact set.
\item Draft questions spanning content and temporal granularities.
\item Enforce the timestamp rules.
\item Retain only unique, answerable questions.
\item Return JSON only.
\end{enumerate}
\end{minipage}}
\caption{Prompt for temporally grounded, open-ended clip-level QA.}
\label{fig:prompt_localqa}
\end{figure}

\subsection{\texorpdfstring{MIT$^3$}{MIT3}-Specific Prompts}

MIT$^3$ introduces two additional prompts for constructing global video knowledge and generating full-video open-ended QA. We provide their complete templates below.
\begin{figure}[t]
\centering
\fcolorbox{black!35}{black!3}{\begin{minipage}{0.92\linewidth}
\textbf{Global memory construction prompt}

\medskip
Create a comprehensive global memory document for the entire video from selected clip descriptions, sorted by time, and a merged entity registry.

\medskip
\textbf{Required sections.}
\begin{enumerate}
\item \textbf{Global entity registry:} disambiguated entities with unique names, appearances, roles, and first/last appearance times.
\item \textbf{Event timeline:} chronological major events with timestamps.
\item \textbf{State changes and causal relationships:} major transitions and their causes.
\item \textbf{Structural overview:} major phases or chapters of the video.
\end{enumerate}
\end{minipage}}
\caption{MIT$^3$ prompt for global memory construction.}
\label{fig:prompt_globalmemory}
\end{figure}

\begin{figure}[t]
\centering
\fcolorbox{black!35}{black!3}{\begin{minipage}{0.92\linewidth}
\textbf{Full-video QA prompt}

\medskip
Generate exactly \texttt{\{N\}} distinct open-ended QA pairs about the entire video.

\medskip
\textbf{Critical disambiguation rules.}
\begin{enumerate}
\item Each question must have a unique, unambiguous answer across the entire video.
\item A question is invalid if multiple scenes or moments could answer it.
\item Questions must not contain timestamps or time references, although answers may mention times.
\item Answers must be definitive rather than speculative.
\item Answers must contain concise factual content (30--100 tokens).
\item Do not generate multiple-choice questions, refusals, or answers requiring knowledge beyond the provided memory.
\end{enumerate}
\end{minipage}}
\caption{MIT$^3$ prompt for open-ended full-video QA.}
\label{fig:prompt_fullqa}
\end{figure}

\section{Additional Implementation Details}
\label{app:implementation_details}

\paragraph{Datasets and Baselines}
LVBench~\citep{wang2025lvbench} contains 103 long videos and 1,549 question--answer pairs covering entity recognition, event understanding, key-information retrieval, temporal grounding, reasoning, and summarization. We report results on both the full benchmark and the subset containing videos longer than one hour. For LongVideoBench~\citep{wu2024longvideobench}, we evaluate on the full benchmark and the LongVideoBench-Long subset adopted by NKR~\citep{guan2026content}, which contains 564 questions over 188 videos of 900--3,600 seconds. Subtitle and audio inputs are excluded from both benchmarks. We additionally evaluate on Video-MME~\citep{fu2025video}, which covers short-, medium-, and long-duration videos across diverse perception and reasoning tasks.

Published baselines include VideoLLaMA3~\citep{zhang2025videollama}, AdaReTaKe~\citep{wang2025adaretake}, VideoRAG~\citep{ren2026videorag}, VCA~\citep{yang2025vca}, Deep Video Discovery (DVD-o3)~\citep{zhang2026deep}, and NKR~\citep{guan2026content}. For direct Qwen3-VL inference, frames are sampled at 2 fps and capped at 768 frames. Since the per-video training data of NKR are unavailable, we follow its data-generation procedure and train the controlled NKR-style baseline and MIT$^3$ using exactly the same generated supervision. The NKR-style baseline optimizes LoRA parameters with only $\mathcal{L}_{\mathrm{NTP}}+\mathcal{L}_{\mathrm{SFT}}$, allowing the effects of causal layer selection and visual memory distillation to be evaluated independently of differences in synthetic supervision.

\paragraph{Optimization Details}
MIT$^3$ is implemented in PyTorch~\citep{paszke2019pytorch}. Following the video-NKR configuration, we use LoRA with rank 256 and scaling factor 512 and optimize the video-specific parameters with AdamW. Unless otherwise specified, each adapter is trained for 5 epochs with a per-device batch size of 4, an initial learning rate of $2\times10^{-4}$ with cosine annealing, and visual-loss weight $\alpha=1$.

For causal localization, we sample at most 20 calibration QA pairs from each temporal scale. For each candidate layer, the intervention gain $G_{i,l}(\lambda)$ is evaluated on
\begin{equation}
    \Lambda = \{0.25, 0.5, 0.75, 1.0\},
\end{equation}
and approximate the equivalent weighted integral in Equation~\ref{eq:memory_utility} by averaging over these four sampled intervention strengths:
\begin{equation}
    R_{i,l}
    \approx
    \frac{1}{|\Lambda|}
    \sum_{\lambda \in \Lambda}
    (1-\lambda)G_{i,l}(\lambda).
\end{equation}
Thus, $b$ serves only as an auxiliary variable in the continuous formulation and is not separately discretized during implementation. For visual memory distillation, we randomly sample 50\% of the temporally grounded QA pairs at each temporal scale and precompute their teacher deltas before adapter optimization. Sensitivity analyses of the top-$K$ layer budget and visual-supervision sampling ratio are provided in Appendix~\ref{app:hyperparameters}.

\paragraph{Training Data}
For a one-hour video, the data-generation pipeline produces approximately 2k multi-scale description examples, 3k temporally grounded local QA pairs, and 500 timestamp-free global QA pairs, yielding approximately 5.5k training examples in total. This corresponds to roughly one-eighth of the training samples used by NKR. To prevent evaluation leakage, we compare the generated supervision against the benchmark evaluation sets and verify that no evaluation questions are included in either adapter training or causal-localization calibration.

We additionally report the per-video construction time measured on four NVIDIA A100 GPUs. For Qwen3-VL-8B, multi-scale data generation, causal localization, visual-supervision generation, and adapter optimization take approximately 30, 20, 10, and 10 minutes, respectively, for approximately 70 minutes in total. For Qwen3-VL-32B, the corresponding stages take approximately 60, 30, 15, and 40 minutes, for approximately 145 minutes in total. Visual-supervision generation includes computing and caching the video-induced teacher deltas used during optimization.

\section{Hyperparameter Analysis}
\label{app:hyperparameters}

We analyze the sensitivity of MIT$^3$ to two key hyperparameters: the number of selected layers $K$ and the sampling ratio used for visual memory distillation. All experiments are evaluated on LVBench, varying one factor at a time while keeping the remaining configuration fixed.

\paragraph{Number of Selected Layers.}
Tables~\ref{tab:k8b} and~\ref{tab:k32b} report the effect of the layer budget $K$ for Qwen3-VL-8B and Qwen3-VL-32B, respectively. For both backbones, performance generally improves as more causally relevant layers are included, but begins to decline once the selected set becomes overly broad. Qwen3-VL-8B achieves its best performance at $K=18$, while Qwen3-VL-32B peaks at $K=32$. This behavior suggests that effective memory internalization benefits from concentrating adaptation on a sufficiently expressive yet selective subset of layers: too few layers provide insufficient capacity for storing video-specific knowledge, whereas expanding the adaptation set beyond the causally salient region may dilute this specialization. We therefore use $K=18$ and $K=32$ for the 8B and 32B models, respectively, throughout the main experiments.

\begin{table}[t]
    \caption{Effect of the selected-layer budget $K$ for Qwen3-VL-8B on LVBench.}
    \label{tab:k8b}
    \centering
    \begin{tabular}{lcccccc}
        \toprule
        $K$ & 1 & 3 & 6 & 12 & 18 & 24 \\
        \midrule
        Accuracy (\%) & 26.1 & 24.9 & 35.6 & 39.6 & \textbf{43.5} & 41.7 \\
        \bottomrule
    \end{tabular}
\end{table}

\begin{table}[t]
    \caption{Effect of the selected-layer budget $K$ for Qwen3-VL-32B on LVBench.}
    \label{tab:k32b}
    \centering
    \begin{tabular}{lccccc}
        \toprule
        $K$ & 12 & 16 & 24 & 32 & 48 \\
        \midrule
        Accuracy (\%) & 37.9 & 42.1 & 46.7 & \textbf{50.6} & 48.1 \\
        \bottomrule
    \end{tabular}
\end{table}

\paragraph{Visual-Supervision Sampling Ratio.}
Table~\ref{tab:visual_sampling} studies the fraction of temporally grounded QA pairs used for visual memory distillation with Qwen3-VL-8B. Removing visual supervision reduces accuracy from 43.5 to 38.2, confirming its importance for effective memory internalization. Increasing the sampling ratio from 50\% to 100\%, however, yields only a marginal improvement of 0.2 points. This saturation indicates that visual memory distillation does not require dense supervision over all local QA pairs; a moderate subset already provides sufficient visual guidance. We therefore use a 50\% sampling ratio in the main experiments to retain nearly the full benefit while reducing the cost of teacher-delta generation.

\begin{table}[t]
    \caption{Effect of the visual-distillation sampling ratio on LVBench with Qwen3-VL-8B.}
    \label{tab:visual_sampling}
    \centering
    \begin{tabular}{cc}
        \toprule
        Visual QA Sampling Ratio & Accuracy (\%) \\
        \midrule
        0\%   & 38.2 \\
        50\%  & 43.5 \\
        100\% & \textbf{43.7} \\
        \bottomrule
    \end{tabular}
\end{table}

\section{Layer-Wise Distribution with Qwen3-VL-32B}
\label{app:layer_distribution_32b}

\begin{figure}[H]
    \centering
    \IfFileExists{figures/layer_distribution_32b.pdf}{
        \includegraphics[width=\linewidth]{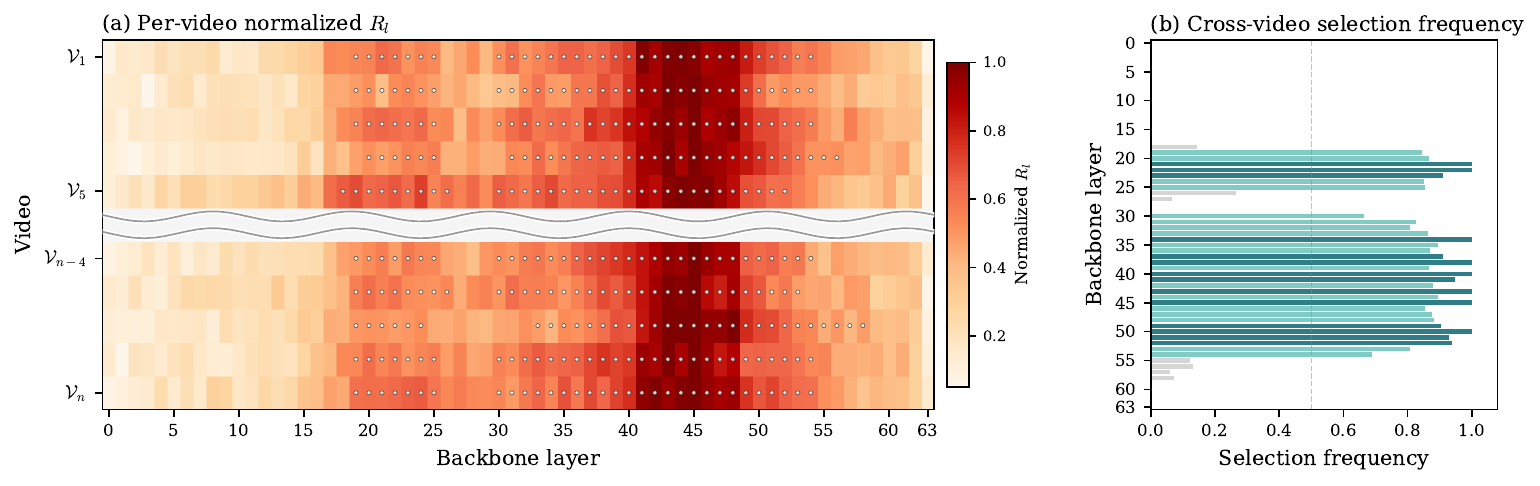}
    }{
        \fbox{\begin{minipage}[c][1.55in][c]{0.94\linewidth}
            \centering
            \textbf{Qwen3-VL-32B layer-distribution visualization}\par\medskip
            \begin{minipage}[t]{0.47\linewidth}
                \centering
                \textbf{(a) Video--layer heatmap}\par
                Rows: videos; columns: backbone layers.\par
                Color: normalized $R_l(V)$; markers: top-32 layers.
            \end{minipage}\hfill
            \begin{minipage}[t]{0.47\linewidth}
                \centering
                \textbf{(b) Cross-video selection frequency}\par
                Horizontal axis: backbone layer.\par
                Vertical axis: fraction of videos selecting the layer.
            \end{minipage}
        \end{minipage}}
    }
    \caption{Layer-wise causal utility and top-32 selection frequency for Qwen3-VL-32B on LVBench.}
    \label{fig:layer_distribution_32b}
\end{figure}

We repeat the layer-distribution analysis with Qwen3-VL-32B, selecting the top-32 adaptation layers for each LVBench video. As in the 8B setting, Figure~\ref{fig:layer_distribution_32b} visualizes the normalized per-video causal utilities together with the cross-video selection frequency. A similar structure emerges at the larger model scale: high-utility layers are concentrated within a shared region across videos, while a smaller subset of selected layers varies with individual video content. The frequently selected layers account for approximately \textbf{72\%} of all top-32 selections, indicating substantial cross-video consistency in the preferred adaptation locations. At the same time, the remaining selections preserve video-specific variation around this shared structure. Overall, the 32B analysis exhibits the same combination of a common memory-writing core and content-dependent refinements observed with Qwen3-VL-8B.

\section{Deployment and Resource Analysis}
\label{app:deployment_cost}

MIT$^3$ offers practical advantages in settings where long videos serve as persistent information sources and are revisited repeatedly. We compare its serving footprint with persistent KV-cache reuse using Qwen3-VL-32B under the same one-hour, 768-frame setting as Table~\ref{tab:main_results}. KV reuse avoids repeated video prefill by retaining the video-conditioned KV cache, whereas MIT$^3$ represents each video with a compact LoRA adapter and performs subsequent inference from text alone.

\begin{table}[H]
\centering
\caption{Persistent-state comparison for Qwen3-VL-32B.}
\label{tab:deployment}
\begin{tabular}{lcc}
\toprule
Setting & KV Reuse & MIT$^3$ \\
\midrule
Per-video persistent state
    & 26.9 GB
    & 0.59 GB \\

Backbone + 1 video
    & 89.0 GB
    & 62.7 GB \\

Backbone + 10 videos
    & 331.1 GB
    & 68.0 GB \\

Reduction vs.\ KV reuse
    & --
    & $45.6\times$ smaller \\
\bottomrule
\end{tabular}
\end{table}

The resource advantage becomes increasingly pronounced as more long videos are retained for subsequent access. Under persistent KV reuse, each additional video introduces a 26.9-GB cache, whereas MIT$^3$ adds only a 0.59-GB adapter. As a result, maintaining ten one-hour videos requires approximately 331 GB including the backbone under KV reuse, compared with only 68 GB for MIT$^3$. This compact per-video state substantially reduces the memory required to maintain and switch among multiple persistent video memories.

Such a deployment pattern is particularly relevant to repeated analysis of long-video collections, including retrospective surveillance review, recorded meetings, lecture archives, and instructional or maintenance videos. In these settings, multiple video memories can share the same backbone and be stored or switched through lightweight adapters, without keeping the original frames or a large video-dependent KV cache resident for every video.

\section{Additional Benchmark Results}
\label{app:additional_results}

\subsection{Video-MME}

Video-MME evaluates general video understanding across short-, medium-, and long-duration videos. Table~\ref{tab:videomme_results} reports overall accuracy with the Qwen3-VL-8B backbone, providing an additional evaluation beyond the dedicated long-video benchmarks. MIT$^3$ consistently improves over the matched NKR-style baseline, while remaining competitive with direct video-conditioned inference despite operating without video input at query time.

\begin{table}[H]
    \caption{Video-MME results. $\dagger$ denotes methods trained with the same data as MIT$^3$.}
    \label{tab:videomme_results}
    \centering
    \begin{tabular}{lcc}
        \toprule
        Method & Input & Accuracy (\%) \\
        \midrule
        Qwen3-VL-8B~\citep{bai2025qwen3} & $<768$ fr. & 69.1 \\
        Qwen3-VL-8B + NKR-style$^{\dagger}$ & N/A & 62.9 \\
        Qwen3-VL-8B + MIT$^3$ (Ours) & N/A & \textbf{66.7} \\
        \bottomrule
    \end{tabular}
\end{table}
\section{Limitations}
\label{app:limitations}

Our study focuses on test-time training, where each video is internalized through lightweight, video-specific adaptation of a pretrained model. This setting allows us to isolate and evaluate the proposed memory-writing mechanisms, but computational constraints limit our exploration of how these mechanisms scale to large-scale pretraining. Extending them to the pretraining stage may support more persistent and reusable video knowledge across videos and tasks, which we leave for future work.

\end{document}